\documentclass[10pt,twocolumn,letterpaper]{article}

\usepackage[pagenumbers]{cvpr} 

\usepackage{multirow}
\usepackage[accsupp]{axessibility}
\definecolor{cvprblue}{rgb}{0.21,0.49,0.74}
\usepackage[pagebackref,breaklinks,colorlinks,allcolors=cvprblue]{hyperref}

\title{Improving Multimodal Hateful Meme Detection Exploiting LMM-Generated Knowledge \thanks{IEEE/CVF CVPR Workshops 2025. Copyright (c) 2025 IEEE. This is the authors' accepted version. The final published article is available in IEEE Xplore.}}

\author{Maria Tzelepi \hspace{8mm} Vasileios Mezaris\\
CERTH-ITI\\
Thessaloniki, Greece\\
{\tt\small \{mtzelepi,bmezaris\}@iti.gr}
}

\begin{document}
\maketitle
\begin{abstract}
Memes have become a dominant form of communication in social media in recent years. Memes are typically humorous and harmless, however there are also memes that promote hate speech, being in this way harmful to individuals and groups based on their identity. Therefore, detecting hateful content in memes has emerged as a task of critical importance. The need for understanding the complex interactions of images and their embedded text renders the hateful meme detection a challenging multimodal task. In this paper we propose to address the aforementioned task leveraging knowledge encoded in powerful Large Multimodal Models (LMM). Specifically, we propose to exploit LMMs in a two-fold manner. First, by extracting knowledge oriented to the hateful meme detection task in order to build strong meme representations. Specifically, generic semantic descriptions and emotions that the images along with their embedded texts elicit are extracted, which are then used to train a simple classification head for hateful meme detection. Second, by developing a novel hard mining approach introducing directly LMM-encoded knowledge to the training process, providing further improvements. We perform extensive experiments on two datasets that validate the effectiveness of the proposed method, achieving state-of-the-art performance.\footnote{Our code and trained models are publicly available at: https://github.com/IDT-ITI/LMM-CLIP-meme.}

\textit{\textcolor{red}{Disclaimer: This paper contains hateful content that may be disturbing to some readers.}}
\end{abstract}    
\section{Introduction}
\label{sec:intro}
Images with embedded text, also known as \textit{memes}, have become a prevalent means of communicating ideas, opinions, emotions, and humor in social media in recent years. Even though their purpose is usually harmless, we are also witnessing their exploitation for promoting hate speech that is harmful against individuals or groups based on specific characteristics, e.g., their race, nationality, religions, gender identity, sexual orientation \cite{kiela2020hateful}. Therefore, detecting hateful memes is a task of critical importance. 

The nature of memes, consisting of images in combination with embedded text that require comprehension of both the involved modalities as well as their interaction, renders hateful meme detection a challenging task. That is, an image and its embedded text may be unimodally harmless but their combination can be harmful \cite{kiela2020hateful}. Earlier works on this task focused on the fusion of the involved modalities using BERT-based models \cite{devlin2019bert}, while more recent methods achieve improved performance by employing powerful CLIP Vision-Language Models (VLM) \cite{radford2021learning}. These CLIP-based methods usually finetune a CLIP model towards the hateful meme detection task, focusing attention on the fusion of the memes' image and text embeddings, since this can be critical considering the aforementioned challenges stemming from the nature of memes. That is, CLIP is trained to maximize the similarity between the image and text embeddings, however this is suitable when image and text pairs bear the same meaning. Contrarily, in memes the image and text may convey contradictory meanings. Moreover, there are some prompt-based approaches that utilize either language models like RoBERTa \cite{liu2019roberta} or large multimodal models (LMM), such as LLaVa \cite{liu2023visual} and GPT-4 \cite{achiam2023gpt}, for predicting whether a meme is hateful. These approaches either use captioners or VLMs, such as CLIP or BLIP-2 \cite{li2023blip}, to extract text-based knowledge, and then propagate it to language models in a unimodal fashion for the hateful meme prediction, or train multimodal language models towards the considered task (particularly there is only one work \cite{lin2024towards} in this direction), rendering them computationally heavy.

\label{sec:proposed}
\begin{figure*}[!h] 
\centerline{\includegraphics[width=0.98\textwidth]{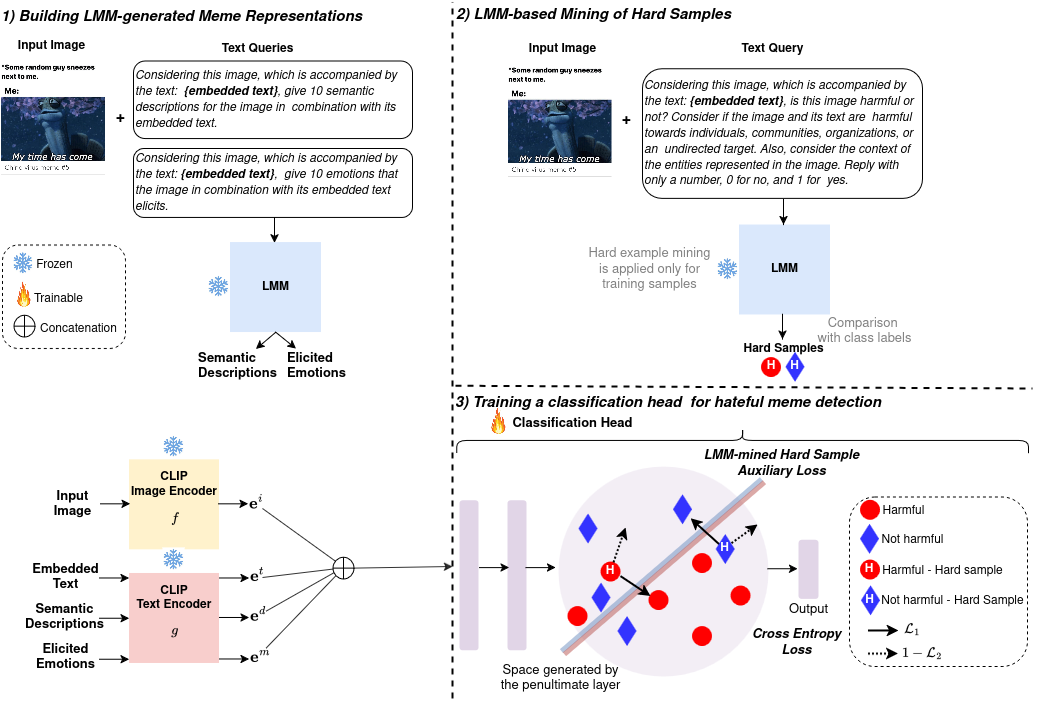}}
\caption{The three steps of the proposed LMM-based training process for hateful meme detection. In the first step we prompt an LMM to extract semantic descriptions and elicited emotions for the memes, and we use a VLM to extract the corresponding embeddings in order to build the meme representations. The meme representations consist of four embeddings: 1) CLIP's image embeddings $\mathbf{e}^i$, 2) CLIP's embedded text embeddings $\mathbf{e}^t$, 3) LMM-generated semantic descriptions embeddings $\mathbf{e}^d$, and 4) LMM-generated elicited emotions embeddings $\mathbf{e}^m$. In the second step we prompt the LMM to identify hard samples. In the third step we train a simple classification head for hateful meme detection using a regular supervised loss and a new LMM-mined hard sample auxiliary loss, using the hard sample information obtained in the second step. At inference time, the first step is executed in order to produce the meme representation for a test meme, which is then is propagated to the trained classification head to predict if it is hateful or not.} 
\label{fig:proposed-method}
\end{figure*}

In this paper, we propose to address the challenging task of hateful meme detection leveraging knowledge encoded in LMMs in a fully multimodal fashion and also using them in an inference only mode (i.e., without training or finetuning LMMs). Specifically, we propose to harness the LMM-encoded knowledge in a two-fold manner (Fig. \ref{fig:proposed-method}): by obtaining knowledge oriented to the considered problem in order to build meaningful meme representations, and by developing a novel LMM-based hard mining approach for further improving the classification performance. To do so, we first prompt a frozen LMM in order to obtain generic semantic descriptions and emotions that the images along with their corresponding embedded texts elicit. Note that this information has been recently exploited for a similar task, i.e., disturbing image detection \cite{tzelepi2024disturbing}, however this is the first time that, oriented to the multimodal hateful meme detection task, such knowledge is exploited in a fully multimodal fashion by prompting the LMM so as to consider the image and its corresponding embedded text. In this way, meaningful information about the memes is extracted, capable of capturing the underlying meanings of the combined modalities. Subsequently, we use a frozen VLM (e.g., CLIP) in order to extract the corresponding embeddings of the LMM-generated knowledge along with the image and text embeddings. We use all the embeddings concatenated as meme representations, in order to train a simple classification head for predicting between hateful and non-hateful memes. 
During the training, apart from the main supervised loss, we propose an LMM-mined hard sample auxiliary loss that forces the hard samples identified by the LMM to become more similar to their nearest non-hard samples that belong to the same class. 

The main contributions of this paper can be summarized as follows:
\begin{itemize}
    \item We leverage LMM-encoded knowledge in a fully multimodal fashion in order to build strong meme representations, that include generic semantic descriptions and elicited emotions, capable of revealing the underlying meanings of the combined modalities and hence addressing the inherent challenges of hateful meme detection.
    \item We additionally use an LMM to identify hard training memes, and propose an LMM-based hard-mining approach that enhances the discrimination ability of the meme embeddings through the LMM-generated hard example information, achieving in turn improved classification performance.
    \item We perform extensive experiments on two challenging datasets in order to validate the effectiveness of the proposed method, achieving state-of-the-art performance.
\end{itemize}

\section{Prior Work}
\label{sec:prior}

Hateful meme detection has become a topic of growing research interest in recent years. Early works for addressing the problem focused on the combination of visual and text features, usually extracted from BERT-based models. For example, DisMultiHate \cite{lee2021disentangling}, aiming to exploit the target entity contextual information (e.g., race, religion), proposed a self-supervised training task, in order to extract the target entities using disentangled latent representations. These representations serve as contextual information to improve hateful meme classification. DisMultiHate used BERT as text encoder to produce latent text representations, while for the visual modality it used the Faster R-CNN \cite{ren2016faster}. In \cite{pramanick2021detecting}, where the Harm-C dataset (with memes related to COVID-19) is introduced, a thorough experimental evaluation of the aforementioned and similar textual, visual and multimodal models is provided. 

More recent methods focus on the utilization of the successful CLIP vision-language model. They often propose finetuning CLIP (with frozen the image and text encoders) by applying projections at the output of the encoders to handle the contradictory meanings of the two modalities and focusing then on fusion of the corresponding outputs. Particularly, the first method to utilize CLIP is MOMENTA \cite{pramanick2021momenta}, which uses CLIP features along with VGG-19 for encoding the ROIs on the visual side, and DistilBERT \cite{sanh2019distilbert} for encoding the topics/entities on the textual side. Subsequently, Hate-clipper \cite{kumar2022hate} proposed the feature interaction matrix (FIM), an intermediate fusion of CLIP's image and text features based on outer product, in order to explicitly model the correlations between the image and text feature spaces. The FIM representation is propagated to a simple classifier for performing hateful meme detection. ISSUES \cite{burbi2023mapping} uses the CLIP model along with the textual inversion technique in order to capture the multimodal semantic content of the memes, for effectively tackling the hateful meme detection task. After training linear projections so as to adapt the embedding spaces of the encoders, it uses a combiner network as a multimodal fusion function whose output is propagated to a multilayer perceptron for the meme classification. SimCLIP \cite{huertas2024clip} proposed to use a pre-trained CLIP encoder in order to produce context-aware embeddings and a Siamese fusion technique in order to capture the interactions between the text and image modalities. Specifically, SimCLIP uses a shallow network to process the image and text embeddings produced by CLIP, then projections are concatenated along with their absolute difference and Hadamard product, and finally the output is propagated to a classification network. Memeclip \cite{shah2024memeclip}, introduced a new dataset, called PrideMM, that contains memes related to the LGBTQ+ Pride movement, and then proposes to adopt lightweight feature adapters for both image and text modalities in order to retain CLIP’s prior knowledge while learning the features of new data. Furthermore, it uses residual connections to integrate prior image and text projections that follow the CLIP's image and text encoders with the outputs of the aforementioned adapters. Contrary to the above CLIP-based approaches, in this paper we propose to capture the complex meaning of the meme's combined modalities, using LMMs to acquire knowledge oriented to the specific task, and we simply use CLIP in order to extract the corresponding embeddings for feeding them to a simple classification head. In this way, CLIP is utilized in a more efficient way, compared to finetuning it with additional projection layers which comes with higher computational cost.

Finally, in the literature there are some works that focus on prompting language models or prompting and training recent state-of-the-art LMMs for hateful meme detection. PromptHate \cite{cao2023prompting} proposed to first convert the multimodal input information into text, using a caption generator and then use this information to prompt a language model for hateful prediction. Specifically, it first constructs meme image captions using ClipCap \cite{mokady2021clipcap} and then prompts, in a unimodal fashion, the pre-trained RoBERTa with two demonstrative examples in order to predict if the input is hateful, leveraging the implicit knowledge in the language model. Similarly, Pro-cap \cite{cao2023pro} proposed to exploit a pre-trained vision-language model in a zero-shot visual question answering  manner for addressing the considered task. Specifically, Pro-cap prompts a frozen BLIP-2 model by asking hateful content-related questions and uses the responses as image captions. ExplainHM \cite{lin2024towards} proposed an explainable approach for hateful meme detection. ExplainHM first conducts a multimodal debate between LMMs in order to produce the explanations obtained from the contradictory harmless and harmful arguments. Then, it finetunes a small language model for classifying memes as harmful or harmless, by exploiting the rationales derived from the contradictory harmfulness arguments as prior knowledge.  Finally, in a similar direction, \cite{zhong2024multimodal} studied the utilization of LMMs (such as LLaVa and MiniGPT-4 \cite{zhu2023minigpt}) in order to produce explanations for memes, concluding that such models can have biases in generating meme explanations, while it also curated a unified dataset for meme explanation. In this paper we also propose a prompt-based methodology using a powerful LMM, however in contrast to existing approaches we use the LMM in a fully multimodal fashion to extract the knowledge (rather than using complex pipelines that employ CLIP-based captioners or VLMs to extract captions and propagating them to language models in a unimodal fashion), directly harnessing the LMM-encoded knowledge, without also depending on the quality of the generated image captions. Moreover, as compared to ExplainHM \cite{lin2024towards}, which is the first work, to the best of our knowledge, that uses LMMs with a different target (i.e., providing explanations), we note that ExplainHM addresses the hateful meme detection task using multimodal language models both for training and inference, contrary to the proposed method that simply extracts the corresponding knowledge and uses it to train a lightweight classification head, rendering the proposed method computationally more efficient.

\section{Proposed Method}\label{sec:proposed}

\subsection{Problem Statement and Method Overview}
In this paper we deal with multimodal hateful meme detection: Given a meme $(\mathbf{I},t)$, consisting of the image $\mathbf{I}$ and the embedded text $t$, we aim to predict whether it is hateful (1) or not (0), i.e, train a classifier  $\mathcal{C}$ such that $\mathcal{C}(\mathbf{I},t) \rightarrow (0,1)$. The proposed method for hateful meme detection, illustrated in Fig. \ref{fig:proposed-method}, consists of three steps. Firstly, we build the meme representations harnessing knowledge tailored to the specific problem from a frozen LMM, and acquire the corresponding embeddings using a frozen CLIP vision-language model. Secondly, we use the LMM in order to obtain its predictions of the training memes being harmful or not, considering in turn the wrong predictions as hard samples. Thirdly, we train a simple classification head for hateful meme detection using as input the meme representations built in the first step, and introducing the LMM-generated hard sample information obtained in the second step to the training process through a new auxiliary hard-mining objective. The latter aims to enhance the discriminability of the aforementioned embeddings, providing further classification improvements.

\subsection{Building LMM-generated Meme Representations}
In the first step, we first use a frozen LMM in order to first obtain generic semantic information, as well as information about the emotions that the memes elicit. To do so, we prompt the LMM in a fully multimodal fashion, giving as input each image of the training dataset along with a text query that contains its embedded text. Note that extracting the embedded text of each meme typically includes an OCR step (several tools can be used for this step, e.g.,\footnote{https://cloud.google.com/vision/docs}), however the utilized datasets already provide this information. The multimodal prompt for the semantic descriptions is articulated as follows: \textit{Considering this image, which is accompanied by the \{embedded text\}, give 10 semantic descriptions for the image in combination with its embedded text.} Similarly, the multimodal prompt for the elicited emotions is articulated as follows: \textit{Considering this image, which is accompanied by the \{embedded text\}, give 10 emotions that the image in combination with its embedded text elicits.} Note that we prompt the LMM for 10 responses, in order to acquire more rich information.

Subsequently, a frozen vision-language model is utilized in order to obtain the image and all the text-based embeddings. Specifically, we use four embeddings: 1) CLIP's image embeddings, 
2) CLIP's embedded text embeddings, 3) LMM-generated semantic descriptions embeddings, and 4) LMM-generated elicited emotions embeddings. 
More specifically, we consider as $\mathcal{I} =\{\mathbf{I}_j \in \mathbb{R}^{h\times w \times c} | j=1, \dots, N\}$ the set of $N$ training images, where $h, w, c$ correspond to the height, width, and channels of the image, respectively. We also consider as $t_j$ the embedded text that accompanies each image $\mathbf{I}_j$, as $d_{j,k}, k=1,\dots,10$ the LMM-generated semantic descriptions for each image, and as $m_{j,k}, k=1,\dots,10$ the LMM-generated elicited emotions for each image. Denoting $\textit{f}$ the CLIP's image encoder, and $\textit{g}$ the CLIP's text encoder, we extract the corresponding image and text embeddings as follows for each image $\mathbf{I}_j$: $\mathbf{e}^i_j = \textit{f}(\mathbf{I}_j) \in \mathbb{R}^{D_1}$, $\mathbf{e}^t_j = \textit{g}(t_j) \in \mathbb{R}^{D_1}$, $\mathbf{e}^d_{j} = avg(\textit{g}(d_{j,k})) \in \mathbb{R}^{D_1}$, and $\mathbf{e}^m_{j} = avg(\textit{g}(m_{j,k})) \in \mathbb{R}^{D_1}$, where $D_1$ is the dimension of the embeddings, and $avg$ is the average pooling operation on the 10 LMM-generated embeddings in order to obtain a unique embedding for each image. Finally, we concatenate all the acquired embeddings, i.e., $[{\mathbf{e}^i_j}^\intercal,{\mathbf{e}^t_j}^\intercal,{\mathbf{e}^d_j}^\intercal,{\mathbf{e}^m_j}^\intercal]^\intercal \in \mathbb{R}^{4 \cdot D_1}$, in order to propagate them as input to the classification head in the third step.

\subsection{LMM-based Mining of Hard Samples}
In the second step, we prompt the LMM for identifying hard samples, in order to exploit them with a new hard-mining auxiliary objective in the third step. Generally, hard example mining is a technique for improving the classification performance focusing on the hard samples, e.g., training samples with the highest loss \cite{shrivastava2016training}. Hard mining has been utilized in the literature on several tasks, and various methods for identifying hard samples have been proposed \cite{yu2018loss,yin2019online,smirnov2018hard}. In this work, given that the meme representations are built mainly bearing knowledge derived from an LMM, we propose to identify the hard training samples by using the same LMM, and directly introduce this knowledge to the training of the classification head. To do so, we directly ask the LMM to classify the training memes as harmful or not, and then comparing the predictions with the actual labels of the memes, we consider the misclassified training samples as hard. The multimodal prompt is articulated as follows: \textit{Considering this image, which is accompanied by the \{embedded text\}, is this image harmful or not? Consider if the image and its text are harmful towards individuals, communities, organizations, or undirected target. Also consider the context of entities represented in the image. Reply with only a number, 0 for no and 1 for yes.}

\subsection{Training a Classification Head for Hateful Meme Detection}
In the third step, we train a simple classification head for discriminating between hateful and not hateful memes. The classification head consists of three trainable linear layers and receives as input the concatenated embedding produced in the first step. Each embedding is associated with a ground-truth class label $l_j \in \{0,1\}$. We train the classification head with a regular supervised loss (i.e., cross entropy loss, $\mathcal{L}_{ce}$) using the class labels, along with an additional \textit{LMM-mined hard sample auxiliary loss}, in order to further leverage knowledge encoded in the LMM. As previously mentioned, we acquire the hard samples in the first step by comparing the LMM-generated predictions with the class labels. Then, considering the embeddings of the feature space generated by its penultimate layer, the additional loss forces the LMM-generated hard embeddings to become more similar to their nearest non-hard embeddings inside the batch that belong to the same class. Note that any layer could be used to apply the additional objective; however, in this work we chose to apply it on the penultimate layer since deeper layers are typically used to apply regularization objectives \cite{tzelepi2020improving}. That is, considering the LMM-mined hard embeddings at the penultimate layer as $\{\mathbf{y}^h_r \in \mathbb{R}^{D_2} | r=1, \dots, N_h\}$, where $N_h$ is the number of hard embeddings and $D_2$ is the dimension of the embeddings, we aim to minimize the Euclidean distance between each hard embedding and the mean vector $\mathbf{m}^r_1 $ of its $n$ nearest non-hard embeddings belonging to the same class. At the same time, in order to ensure that the additional objective will prevent the embeddings entanglement, we also force each hard embedding to move away from its nearest embeddings of the opposite class, i.e., we aim to maximize the Euclidean distance between each hard embedding, $\mathbf{y}^h_r$ and the mean vector $\mathbf{m}^r_2 $ of its $n$ nearest embeddings belonging to the opposite class. Note that the additional computational cost for computing the nearest embeddings inside the batch is manageable, given the typical batch size values; however, it could be further  minimized by incorporating approximate nearest neighbors techniques. The additional auxiliary objectives are formulated as follows: 
\begin{equation}
    \mathcal{L}_{1}= \sum_{r=1}^{N_h} \lVert\mathbf{y}^h_r-\mathbf{m}_1^r \rVert_2^2,
\end{equation}
and
\begin{equation}
    \mathcal{L}_{2}= \sum_{r=1}^{N_h} \lVert\mathbf{y}^h_r-\mathbf{m}_2^r \rVert_2^2.
\end{equation}
Thus, the overall LMM-mined hard sample auxiliary loss, $\mathcal{L}_{HM}$, is formulated as:  $\mathcal{L}_{HM} = \mathcal{L}_{1} + (1-\mathcal{L}_{2})$, while the total loss,  $\mathcal{L}_{total}$, is formulated as: 
\begin{equation}\label{eq:loss}
    \mathcal{L}_{total}=\mathcal{L}_{ce} + \alpha\mathcal{L}_{HM},
\end{equation}
where the parameter $\alpha$ controls the contribution of the two loss terms. In this way, the classification head is trained for discriminating between hateful and not hateful samples due to the main classification loss, and at the same time improved generalization ability is achieved exploiting directly LMM-encoded knowledge, leading to advanced classification performance. It should be finally highlighted that the proposed LMM-based hard mining approach is applied only on the training samples during the training process. During the test process, the test meme representations are built based on the first step of the proposed method, and the test samples are propagated to the trained classification head in order to predict if they are hateful or not.

\section{Experimental Evaluation}
\label{sec:exps}

\subsection{Datasets and Evaluation Metrics}
We evaluate the effectiveness of the proposed methodology for hateful meme detection on two datasets: Harm-C \cite{pramanick2021detecting} and PrideMM \cite{shah2024memeclip}. Harm-C consists of text-embedded images, associated with COVID-19, labeled with three classes: \textit{very harmful}, \textit{partially harmful}, and \textit{harmless}. Similarly to the majority of previous works (e.g., \cite{lin2024towards,cao2023prompting,burbi2023mapping}) we combine the first two classes into \text{harmful}, forming a binary classification problem. Harm-C consists of 3,013 training, 531 validation, and 354 test images. PrideMM is a very recent dataset consisting of text-embedded images, associated with LGBTQ+ Pride movement, labeled as hateful and non hateful. The dataset consists of 4,017 training, 213 validation, and 472 test images (note that 361 images of the provided dataset are corrupted and could not be used).

We evaluate the performance in terms of accuracy similarly to previous works (e.g., \cite{kumar2022hate,pramanick2021momenta,lin2024towards}) i.e., the ratio of number of correct predictions to the total number of input samples. We perform each experiment five times and we report the mean value and the standard deviation. 

\begin{table*}
    \centering
    \caption{LMM-based hard samples from the Harm-C (left) and PrideMM (right) datasets.}
    \label{tab:hard}
    \resizebox{0.98\textwidth}{!}{
    \begin{tabular}{|c|c|c|c||c|c|c|c|}
        \hline
          &   &  \textbf{Ground-} & \textbf{LMM} &   &    &  \textbf{Ground} & \textbf{LMM} \\
        \textbf{Meme}  &  \textbf{Embedded Text}  &  \textbf{Truth} & \textbf{Prediction} & \textbf{Meme}  &  \textbf{Embedded Text}  &  \textbf{Truth} & \textbf{Prediction} \\       
        \hline
        \parbox{2.8cm}{\includegraphics[width=1.2in]{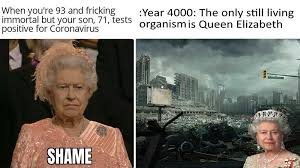}} & \parbox{5cm}{When you're 93 and fricking immortal but your son, 71, tests organism is Queen Elizabeth positive for Córonavirus:Year 4000: The only still living SHAME} & \parbox{1.3cm}{Not Harmful} & \parbox{1.3cm}{Harmful} & \parbox{2.8cm}{\includegraphics[width=1.2in]{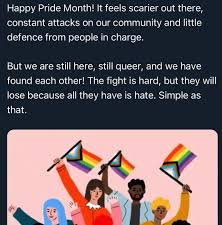}} & \parbox{5cm}{happy pride month it feels scarier out there constant attacks on our community and little defence from people in charge but we are still here still queer and we have found each other the fight is hard but they will lose because all they have is hate simple as that } & \parbox{1.3cm}{Not Harmful} & \parbox{1.3cm}{Harmful} \\ \hline
        \parbox{2.8cm}{\includegraphics[width=1.2in]{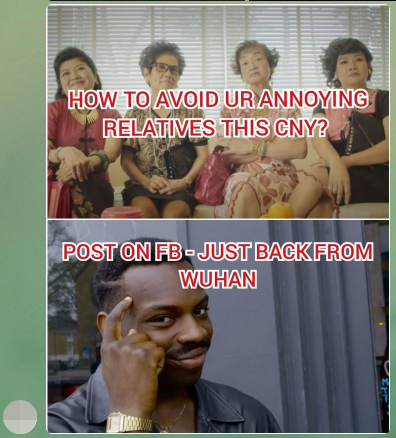}} & \parbox{5cm}{HOW TO AVOID UR ANNOYING RELATIVES THIS CNY? POST ON FB-JUST BACKFROM WUHAN} & \parbox{1.3cm}{Harmful} & \parbox{1.3cm}{Not Harmful} &
        \parbox{2.8cm}{\includegraphics[width=1.2in]{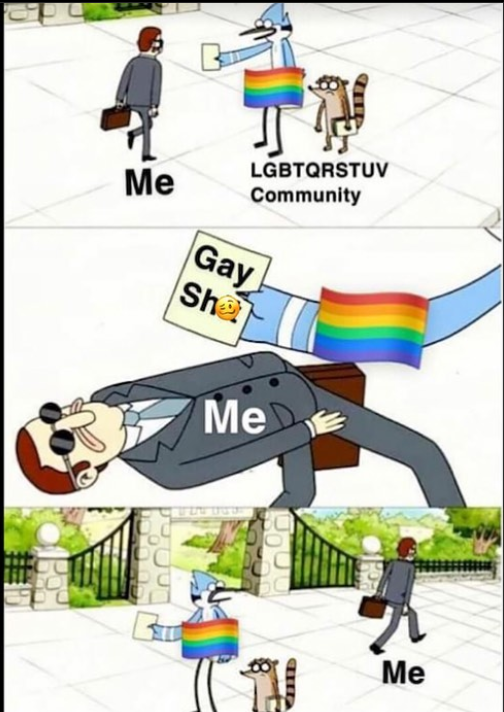}} & \parbox{5cm}{me th lgbtqrstuv community gay she me me} &  \parbox{1.3cm}{Harmful} & \parbox{1.3cm}{Not Harmful} \\ \hline
    \end{tabular}}
\end{table*}

\begin{table*}
    \centering
    \caption{Qualitative results on the LMM-generated semantic descriptions and elicited emotions.}
     \resizebox{0.99\textwidth}{!}{
    \label{tab:qual}
    \begin{tabular}{|c|c|c|}
        \hline
        \textbf{Meme}  &  \textbf{Semantic Descriptions}  &  \textbf{Elicited Emotions} \\
        \hline
        \parbox{4cm}{\includegraphics[width=1.5in]{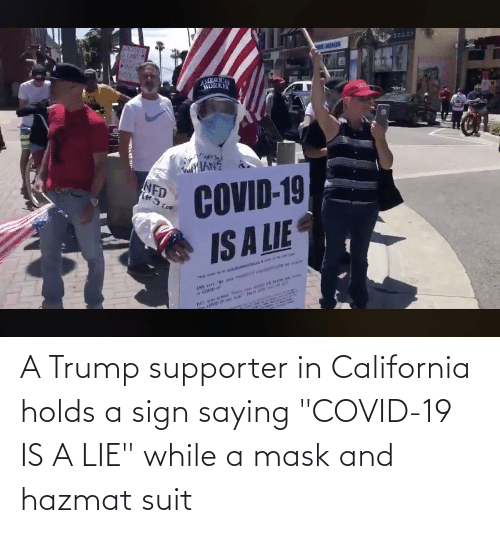}} & \parbox{10cm}{\small1. A person holding a sign that reads "COVID-19 IS A LIE" while wearing a mask and hazmat suit.
2. A protester in California holding a sign with a political message.
3. A person demonstrating against COVID-19, wearing protective gear.
4. A political activist holding a sign with a controversial statement.
5. A person standing on the street with a sign expressing their opinion on COVID-19. 6. A protester holding a sign with a message about COVID-19 in California. 7. A person wearing a mask and hazmat suit, holding a sign with a political statement. 8. A demonstrator standing on the street with a sign and protective gear. 9. A political activist holding a sign with a controversial message about COVID-19. 
} & \parbox{2.5cm}{\small 1. Anger\\
2. Disbelief\\
3. Frustration\\
4. Outrage\\
5. Disapproval\\
6. Disgust\\
7. Sadness\\
8. Fear\\
9. Concern\\
10. Sympathy}   \\ \hline
        \parbox{4cm}{\includegraphics[width=1.6in]{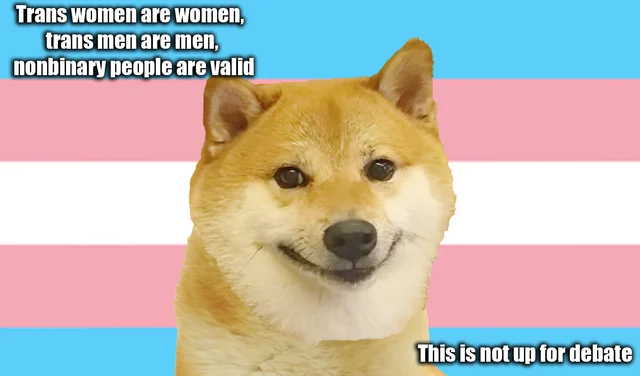}} & \parbox{10cm}{\small 1. The image depicts a dog with a caption that reads, "Trans women are women, trans men are men, nonbinary people are valid. This is not up for debate."
2. The image shows a cute dog with a message that advocates for the acceptance of trans women and trans men as valid members of society.
3. The image features a dog with a caption that emphasizes the importance of recognizing nonbinary people as valid individuals.
4. The image displays a dog with a message that promotes the idea that trans women and trans men are not up for debate.
5. The image shows a dog with a caption that stresses the need to validate nonbinary people and their identities.
6. The image displays a dog with a message that advocates for the recognition of trans women and trans men as valid individuals.
} & \parbox{2.5cm}{\small 1. Confidence\\ 2. Empowerment\\ 3. Pride\\ 4. Strength\\ 5. Courage\\ 6. Resilience\\ 7. Determination\\ 8. Assertiveness\\ 9. Self-esteem\\10. Validity}   \\ \hline
        \parbox{4cm}{\includegraphics[width=1.6in]{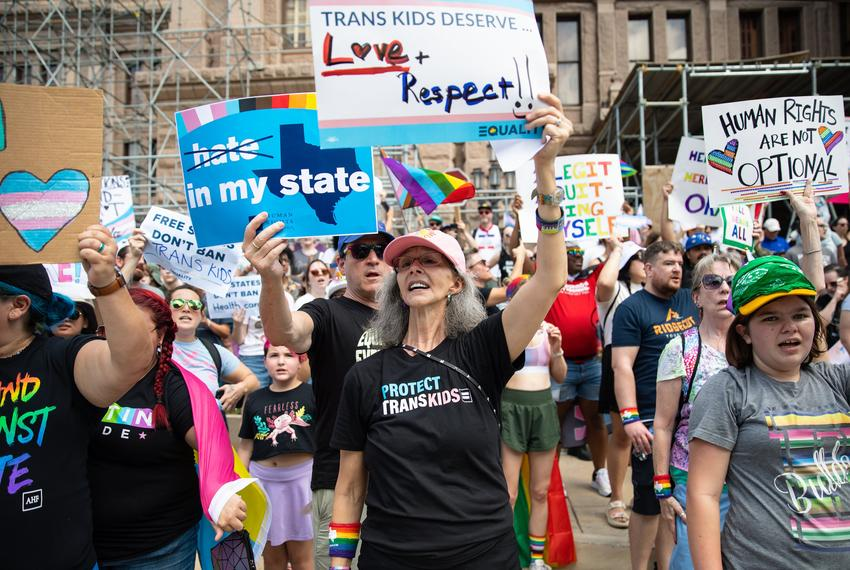}}& \parbox{10cm}{\small 1. A group of people holding up signs in support of transgender rights.
2. A crowd gathered in front of a building holding up signs and banners.
3. A demonstration in support of transgender equality with people holding up signs and banners.
4. A group of people holding up signs and banners in support of transgender rights.
5. A crowd gathered in front of a building holding up signs and banners in support of transgender equality.
6. A demonstration in support of transgender rights with people holding up signs and banners.
7. A group of people holding up signs and banners in support of transgender equality.
8. A crowd gathered in front of a building holding up signs and banners in support of transgender rights.
} & \parbox{2.5cm}{\small 1. Joy\\ 2. Pride\\ 3. Hope\\ 4. Empathy\\ 5. Understanding\\ 6. Acceptance\\ 7. Love\\ 8. Respect \\9. Protection\\ 10. Inclusion}   \\ \hline
    \end{tabular}}
\end{table*}

\subsection{Experimental Setup}
In this paper we use MiniGPT-4 \cite{zhu2023minigpt} for obtaining the LMM-based knowledge which is an open-source LMM that aligns a frozen visual encoder (ViT-G/14 \cite{zhai2022scaling}) with a frozen LLM (Vicuna \cite{vicuna2023}), using a single projection layer. Concerning the VLM for extracting the image and text embeddings, we experiment with ViT-L-14 CLIP \cite{radford2021learning} and LongCLIP-L \cite{zhang2024long}. The classification head for classifying between harmful and non-harmful images consists of three linear layers of 512, 256 and 2 neurons (output). The classification head is trained for 500 epochs, the learning rate is set to 0.001, the parameter $\alpha$ in Eq. (\ref{eq:loss}) is set to 0.05, and the batch size is set to 64 samples. Experiments are conducted on an NVIDIA GeForce RTX 4090 with 24 GB of GPU memory.

\subsection{Experimental Results}
First, regarding the hard mining step of the proposed method, the hard samples obtained from the LMM are 1,089 out of 3,013 training samples of the Harm-C dataset, while for the PrideMM dataset, the LMM produced 1,540 hard samples. This also highlights the ineffectiveness of simply using the LMM as a zero-shot classifier for the hateful meme detection task. Some examples of the obtained hard samples, including not harmful memes that were classified by the LMM as harmful, as well as harmful memes that were classified as not harmful, for both the utilized datasets are provided in Table \ref{tab:hard}. We can observe that the misclassified memes are usually \textit{marginal} cases, for example in the Harm-C dataset considering the misclassified as not harmful example, the meme is annotated as \textit{somewhat harmful} in the original dataset (note that we used the binary version of the dataset that considers as harmful both the somewhat and very harmful memes). Additionally, we can notice in the example of the PrideMM sample misclassified as harmful, that possibly the LMM focused on the embedded text specific words, such as \textit{feels scarier, attacks on our community}, failing to capture the positive message of the meme.

In Table \ref{tab:qual} we provide some qualitative results for demonstrating the meaningfulness of the LMM-generated responses used for building the meme representations, considering three memes from the Harm-C and PrideMM datasets (in interest of space, some of the 10 semantic descriptions per meme have been omitted). The first meme is \textit{somewhat harmful}, while the rest two are harmless and positive. As demonstrated, the semantic descriptions as well as the elicited emotions provide rich information which is complementary to the meme's original information as well as to each other, achieving in this way to capture the underlying meaning and convey their somewhat negative and unsettling message in the first case and the positive message in the latter two.

In Table \ref{tab:sota} we provide a comparison of the proposed method against state-of-the-art methods for hateful meme detection. As mentioned previously, we use both CLIP and LongCLIP models. It should be emphasized that the recent LongCLIP accepts input up to 248 tokens (while the corresponding limit for the CLIP model is 77 tokens), and provides improved performance as demonstrated in \cite{zhang2024long}. The utilization of LongCLIP is grounded in this attribute of longer input limit, since it can be beneficial considering the often long embedded text of the memes. We also note that in the provided results we have used the one-nearest non-hard embedding in the LMM-based auxiliary objective. In the ablation study we experiment with different numbers of nearest embeddings. We compare the proposed method both with CLIP-based methods (\cite{kumar2022hate,burbi2023mapping,shah2024memeclip,pramanick2021momenta}) as well as with prompt-based methods (\cite{cao2023prompting,cao2023pro}). Additionally, we compare the proposed method with \cite{lin2024towards} that involves training of LLMs (marked with $^\star$ in the Table), even though in our approach we only use an LMM in inference mode, and we simply train a classification head for addressing the task. That is, the proposed method is significantly more computationally efficient compared to \cite{lin2024towards}. Best performance on each dataset is printed bold. As demonstrated, the proposed method provides state-of-the-art performance on both the utilized datasets, in the Harm-C using the LongCLIP model, while in the PrideMM using the CLIP model. Note that the behavior of the proposed method on PrideMM dataset with respect to the CLIP and LongCLIP models is extensively discussed in the ablation study. 

\begin{table}
\begin{center}
\caption{Comparisons with state-of-the-art in terms of accuracy (\%). The results are in the form of mean $\pm$ standard deviation across five runs. A * indicates a method that involves training of an LLM. The experimental results are as presented in the corresponding paper, except for result marked with $\diamond$, where the results are reported according to \cite{shah2024memeclip} across three runs.} \label{tab:sota}
\resizebox{0.46\textwidth}{!}{
\begin{tabular}{|c|c|c|}
  \hline
  \bf{Method} &  \bf{Harm-C} &  \bf{PrideMM}
  \\
  \hline
  MOMENTA$^{\diamond}$ \cite{pramanick2021momenta} \small \textcolor{gray}{(EMNLP '21)} & 82.44 $\pm$ 0.65 & 72.23 $\pm$ 0.58\\ \hline 
  DisMultiHate \cite{lee2021disentangling} \small \textcolor{gray}{(ACMMM '21)} & 81.24 $\pm$ 1.04 & - \\ \hline  
  Hate-CLIPper$^{\diamond}$ \cite{kumar2022hate} \small \textcolor{gray}{(NLP4PI '22)} &  83.68 $\pm$ 0.62& 75.53 $\pm$ 0.58\\ \hline 
  PromptHate \cite{cao2023prompting} \small \textcolor{gray}{(EMNLP '22)} & 84.47 $\pm$ 1.75 & - \\ \hline  
  ISSUES$^{\diamond}$ \cite{burbi2023mapping} \small \textcolor{gray}{(CVPR '23)} & 81.31 $\pm$ 1.05 & 74.68 $\pm$ 1.62\\ \hline
  Pro-Cap \cite{cao2023pro} \small \textcolor{gray}{(ACMMM '23)} & 85.03 $\pm$ 1.51 & - \\ \hline 
  MemeCLIP \cite{shah2024memeclip} \small \textcolor{gray}{(EMNLP '24)} & 84.72 $\pm$ 0.45 & 76.06 $\pm$ 0.23\\ \hline 
  ExplainHM$^\star$ \cite{lin2024towards} \small \textcolor{gray}{(WWW '24)} & 87.00  & - \\ \hline \hline
  LMM-CLIP (Proposed) & 86.33 $\pm$ 0.42 & \bf{76.31 $\pm$ 0.39}\\ \hline 
  LMM-LongCLIP (Proposed) & \bf{87.23 $\pm$ 0.33} & 75.89 $\pm$ 0.54\\ \hline 
\end{tabular}}
\end{center}
\end{table}

\begin{table}
\begin{center}
\caption{Evaluation of the proposed method in terms of accuracy (\%) for different numbers of nearest embeddings ($n$) in the LMM-mined hard sample auxiliary loss, using CLIP and LongCLIP models, on the Harm-C and PrideMM datasets. The results are in the form of mean $\pm$ standard deviation across five runs.} \label{tab:ne}
\resizebox{0.46\textwidth}{!}{%
\begin{tabular}{|c|c|c||c|c|c|}
  \hline
 \multirow{2}{*}{\bf{$n$}}  & \multicolumn{2}{c||}{\bf{Harm-C}} & \multicolumn{2}{c|}{\bf{PrideMM}} \\ \cline{2-5}
 & \bf{CLIP} &  \bf{LongCLIP} & \bf{CLIP} &  \bf{LongCLIP}
  \\ \hline
  0  & 85.65 $\pm$ 0.42 & 86.21 $\pm$ 0.21 & 75.89 $\pm$ 0.53 & 75.47 $\pm$ 0.47 \\ \hline 
   1  & 86.33 $\pm$ 0.42 & \bf{87.23 $\pm$ 0.33}  & \bf{76.31 $\pm$ 0.39} & \bf{75.89 $\pm$ 0.54}\\ \hline 
   2  & \bf{86.44 $\pm$ 0.47}  & 86.95 $\pm$ 0.55 & 76.10 $\pm$ 0.25 & 75.68 $\pm$ 0.16 \\ \hline 
   4  &  86.10 $\pm$ 0.33 & 86.78 $\pm$ 0.33 &  76.14 $\pm$ 0.29 & 75.80 $\pm$ 0.36\\ 
  \hline 
\end{tabular}}
\end{center}
\end{table}

\begin{table*}
\begin{center}
\caption{Ablations in terms of accuracy (\%) using CLIP and LongCLIP models on Harm-C and PrideMM datasets. The results are in the form of mean $\pm$ standard deviation across five runs.} \label{tab:abl}
\resizebox{0.98\textwidth}{!}{%
\begin{tabular}{|c|c|c|c|c|c|c||c|c|}
  \hline
\multicolumn{5}{|c|}{\bf{Embeddings}} & \multicolumn{2}{c||}{\bf{Harm-C}} & \multicolumn{2}{c|}{\bf{PrideMM}} \\ \hline
 \bf{Image} & \bf{Embedded Text} & \bf{Semantic Descriptions} & \bf{Elicited Emotions} & \bf{HM}&  \bf{CLIP} &  \bf{LongCLIP} & \bf{CLIP} &  \bf{LongCLIP}
  \\ \hline
 \checkmark & \checkmark & & & & 84.63 $\pm$ 0.29  & 85.59 $\pm$ 0.25 &  75.21 $\pm$ 0.27    & 76.06 $\pm$ 0.27\\ \hline 
 \checkmark & \checkmark & \checkmark & & & 85.25 $\pm$ 0.49 & 86.05 $\pm$ 0.38  & 75.51 $\pm$ 0.29  & 75.38 $\pm$ 0.71\\ \hline 
 \checkmark & \checkmark & & \checkmark & & 85.31 $\pm$ 0.36 & 85.99 $\pm$ 0.23 & 75.59 $\pm$ 0.25  & 76.02 $\pm$ 0.41\\ \hline 
 \checkmark & \checkmark & & & \checkmark & 85.20 $\pm$ 0.46 & 86.16 $\pm$ 0.74 & 75.76 $\pm$ 0.39  & \bf{76.48 $\pm$ 0.35}\\ \hline 
  \checkmark & \checkmark & \checkmark & \checkmark & & 85.65 $\pm$ 0.42 & 86.21 $\pm$ 0.21 & 75.89 $\pm$ 0.53 & 75.47 $\pm$ 0.47\\ \hline
  \checkmark & \checkmark & \checkmark &\checkmark  & \checkmark & \bf{86.33 $\pm$ 0.42} & \bf{87.23 $\pm$ 0.33} &\bf{76.31 $\pm$ 0.39}   & 75.89 $\pm$ 0.54\\ \hline  
\end{tabular}}
\end{center}
\end{table*}

\subsection{Ablation Study}

In Table \ref{tab:ne} we present the experimental results for the proposed method on both Harm-C and PrideMM datasets using CLIP and LongCLIP models, for different numbers of nearest embeddings, $n$ in the LMM-mined hard sample auxiliary loss, against baseline. As baseline we consider the proposed method without applying additional hard-mining objective (i.e., $n$=0). That is, we train the classification head using the same input representation (image, embedded text, semantic descriptions and elicited emotions) without the auxiliary objective. Best results for each dataset and VLM are printed in bold. 
From the demonstrated results we can observe that the proposed additional hard mining objective significantly improves the baseline performance in all the considered cases, for all the considered numbers of nearest embeddings. Furthermore, we can observe that better performance is obtained for fewer nearest embeddings (in three out of four cases using only the first nearest embedding, while in one out of three using the first two nearest embeddings). 

Subsequently, we perform experiments using each of the proposed components separately. The results are provided in Table \ref{tab:abl}. As it can be observed, in all the considered cases, apart from the one of LongCLIP on PrideMM, each of the proposed components improves the baseline performance, while the combination of all the proposed components provides further improvements. Considering the case of LongCLIP on PrideMM dataset, we discovered in the conducted ablation that the semantic descriptions embeddings and marginally the elicited emotion embeddings harm the baseline performance. This is attributed to the fact that this information can be redundant when the embedded text information is rich enough, and it can be exploited, e.g., using LongCLIP. Specifically, we observed that the average number of tokens of the embedded texts in PrideMM is considerably larger than the one in Harm-C, i.e., 42.66 against 26.52. Moreover, the embedded text of only 64 images is over 77 tokens in the Harm-C dataset, while in PrideMM there are 489 images with embedded text over 77 tokens. Note that the CLIP's text encoder truncates the text to the first 77 tokens, discarding the rest. Thus, in PrideMM, optimally exploiting the embedded text information, using LongCLIP, renders the semantic descriptions and elicited emotions knowledge redundant. Hence, even though the proposed LMM-mined hard sample auxiliary loss improves the performance of using all the proposed input embeddings, the final performance of the proposed method is lower than the baseline, due the decrease of mainly the semantic descriptions input. In this case, the best performance is obtained only with the proposed LMM-mined hard sample auxiliary loss using the baseline embeddings as input.

\section{Conclusions}
\label{sec:con}

In this paper we proposed an LMM-based method for dealing with the hateful meme detection task. The proposed method first prompts the LMM in a fully multimodal fashion in order to obtain knowledge oriented to the specific task for building strong meme representations that can address the inherent challenges associated with the considered task. A CLIP-based model is used to extract the corresponding embeddings in order to train then a simple classification head for hateful meme detection. Second, the LMM is prompted in order to generate predictions for the training memes. Then it uses the predictions in order to obtain hard samples, and introduces the LMM-generated hard sample information directly to the training process of the classification head, through an additional auxiliary loss. Training the classifier with both the main classification and the auxiliary objective achieves improved classification performance. The effectiveness of the proposed method is validated through extensive experiments on two datasets, achieving state-of-the-art performance. 

\textbf{Limitations}: From an ethical viewpoint, the nature of hateful meme detection includes a certain level of subjectivity, rendering it difficult to clearly distinguish between offensive and not offensive content.
Apart from the general limitations, even though the proposed method is computationally more efficient compared to methods that train/finetune multimodal language models, it still uses an LMM for inference, which still comes with a certain computational cost. Experimenting with smaller LMMs would make the proposed method easier to deploy on resource-constrained devices. Furthermore, the application of the proposed method on multilingual memes is linked with the availability and effectiveness of multilingual vision language models.


{
    \small

\begin{thebibliography}{31}
\providecommand{\natexlab}[1]{#1}
\providecommand{\url}[1]{\texttt{#1}}
\expandafter\ifx\csname urlstyle\endcsname\relax
  \providecommand{\doi}[1]{doi: #1}\else
  \providecommand{\doi}{doi: \begingroup \urlstyle{rm}\Url}\fi

\bibitem[Achiam et~al.(2023)Achiam, Adler, Agarwal, Ahmad, Akkaya, Aleman, Almeida, Altenschmidt, Altman, Anadkat, et~al.]{achiam2023gpt}
Josh Achiam, Steven Adler, Sandhini Agarwal, Lama Ahmad, Ilge Akkaya, Florencia~Leoni Aleman, Diogo Almeida, Janko Altenschmidt, Sam Altman, Shyamal Anadkat, et~al.
\newblock Gpt-4 technical report.
\newblock \emph{arXiv preprint arXiv:2303.08774}, 2023.

\bibitem[Burbi et~al.(2023)Burbi, Baldrati, Agnolucci, Bertini, and Del~Bimbo]{burbi2023mapping}
Giovanni Burbi, Alberto Baldrati, Lorenzo Agnolucci, Marco Bertini, and Alberto Del~Bimbo.
\newblock Mapping memes to words for multimodal hateful meme classification.
\newblock In \emph{Proceedings of the IEEE/CVF International Conference on Computer Vision}, pages 2832--2836, 2023.

\bibitem[Cao et~al.(2022)Cao, Lee, Chong, and Jiang]{cao2023prompting}
Rui Cao, Roy Ka-Wei Lee, Wen-Haw Chong, and Jing Jiang.
\newblock Prompting for multimodal hateful meme classification.
\newblock In \emph{Proceedings of the Conference on Empirical Methods in Natural Language Processing}, pages 321--332, 2022.

\bibitem[Cao et~al.(2023)Cao, Hee, Kuek, Chong, Lee, and Jiang]{cao2023pro}
Rui Cao, Ming~Shan Hee, Adriel Kuek, Wen-Haw Chong, Roy Ka-Wei Lee, and Jing Jiang.
\newblock Pro-cap: Leveraging a frozen vision-language model for hateful meme detection.
\newblock In \emph{Proceedings of the ACM International Conference on Multimedia}, pages 5244--5252, 2023.

\bibitem[Chiang et~al.(2023)Chiang, Li, Lin, Sheng, Wu, Zhang, Zheng, Zhuang, Zhuang, Gonzalez, Stoica, and Xing]{vicuna2023}
Wei-Lin Chiang, Zhuohan Li, Zi Lin, Ying Sheng, Zhanghao Wu, Hao Zhang, Lianmin Zheng, Siyuan Zhuang, Yonghao Zhuang, Joseph~E. Gonzalez, Ion Stoica, and Eric~P. Xing.
\newblock Vicuna: An open-source chatbot impressing gpt-4 with 90\%* chatgpt quality, 2023.

\bibitem[Devlin et~al.(2019)Devlin, Chang, Lee, and Toutanova]{devlin2019bert}
Jacob Devlin, Ming-Wei Chang, Kenton Lee, and Kristina Toutanova.
\newblock Bert: Pre-training of deep bidirectional transformers for language understanding.
\newblock In \emph{Proceedings of the North American Chapter of the Association for Computational Linguistics: Human Language Technologies}, pages 4171--4186, 2019.

\bibitem[Huertas-Tato et~al.(2024)Huertas-Tato, Koutlis, Papadopoulos, Camacho, and Kompatsiaris]{huertas2024clip}
Javier Huertas-Tato, Christos Koutlis, Symeon Papadopoulos, David Camacho, and Ioannis Kompatsiaris.
\newblock A clip-based siamese approach for meme classification.
\newblock In \emph{Proceedings of the International Joint Conference on Neural Networks}, pages 1--8. IEEE, 2024.

\bibitem[Kiela et~al.(2020)Kiela, Firooz, Mohan, Goswami, Singh, Ringshia, and Testuggine]{kiela2020hateful}
Douwe Kiela, Hamed Firooz, Aravind Mohan, Vedanuj Goswami, Amanpreet Singh, Pratik Ringshia, and Davide Testuggine.
\newblock The hateful memes challenge: Detecting hate speech in multimodal memes.
\newblock \emph{Proceedings of Advances in Neural Information Processing Systems}, 33:\penalty0 2611--2624, 2020.

\bibitem[Kumar and Nandakumar(2022)]{kumar2022hate}
Gokul~Karthik Kumar and Karthik Nandakumar.
\newblock Hate-clipper: Multimodal hateful meme classification based on cross-modal interaction of clip features.
\newblock In \emph{Proceedings of the NLP4PI Workshop}, pages 171--183, 2022.

\bibitem[Lee et~al.(2021)Lee, Cao, Fan, Jiang, and Chong]{lee2021disentangling}
Roy Ka-Wei Lee, Rui Cao, Ziqing Fan, Jing Jiang, and Wen-Haw Chong.
\newblock Disentangling hate in online memes.
\newblock In \emph{Proceedings of the ACM International Conference on Multimedia}, pages 5138--5147, 2021.

\bibitem[Li et~al.(2023)Li, Li, Savarese, and Hoi]{li2023blip}
Junnan Li, Dongxu Li, Silvio Savarese, and Steven Hoi.
\newblock Blip-2: Bootstrapping language-image pre-training with frozen image encoders and large language models.
\newblock In \emph{Proceedings of International Conference on Machine Learning}, pages 19730--19742. PMLR, 2023.

\bibitem[Lin et~al.(2024)Lin, Luo, Gao, Ma, Wang, and Yang]{lin2024towards}
Hongzhan Lin, Ziyang Luo, Wei Gao, Jing Ma, Bo Wang, and Ruichao Yang.
\newblock Towards explainable harmful meme detection through multimodal debate between large language models.
\newblock In \emph{Proceedings of the ACM on Web Conference}, pages 2359--2370, 2024.

\bibitem[Liu et~al.(2023)Liu, Li, Wu, and Lee]{liu2023visual}
Haotian Liu, Chunyuan Li, Qingyang Wu, and Yong~Jae Lee.
\newblock Visual instruction tuning.
\newblock In \emph{Proceedings of Advances in Neural Information Processing Systems}, pages 34892--34916, 2023.

\bibitem[Liu et~al.(2019)Liu, Ott, Goyal, Du, Joshi, Chen, Levy, Lewis, Zettlemoyer, and Stoyanov]{liu2019roberta}
Yinhan Liu, Myle Ott, Naman Goyal, Jingfei Du, Mandar Joshi, Danqi Chen, Omer Levy, Mike Lewis, Luke Zettlemoyer, and Veselin Stoyanov.
\newblock Roberta: A robustly optimized bert pretraining approach.
\newblock \emph{arXiv preprint arXiv:1907.11692}, 2019.

\bibitem[Mokady et~al.(2021)Mokady, Hertz, and Bermano]{mokady2021clipcap}
Ron Mokady, Amir Hertz, and Amit~H Bermano.
\newblock Clipcap: Clip prefix for image captioning.
\newblock \emph{arXiv preprint arXiv:2111.09734}, 2021.

\bibitem[Pramanick et~al.(2021{\natexlab{a}})Pramanick, Dimitrov, Mukherjee, Sharma, Akhtar, Nakov, and Chakraborty]{pramanick2021detecting}
Shraman Pramanick, Dimitar Dimitrov, Rituparna Mukherjee, Shivam Sharma, Md~Shad Akhtar, Preslav Nakov, and Tanmoy Chakraborty.
\newblock Detecting harmful memes and their targets.
\newblock \emph{arXiv preprint arXiv:2110.00413}, 2021{\natexlab{a}}.

\bibitem[Pramanick et~al.(2021{\natexlab{b}})Pramanick, Sharma, Dimitrov, Akhtar, Nakov, and Chakraborty]{pramanick2021momenta}
Shraman Pramanick, Shivam Sharma, Dimitar Dimitrov, Md~Shad Akhtar, Preslav Nakov, and Tanmoy Chakraborty.
\newblock Momenta: A multimodal framework for detecting harmful memes and their targets.
\newblock In \emph{Proceedings of Conference on Empirical Methods in Natural Language Processing}, 2021{\natexlab{b}}.

\bibitem[Radford et~al.(2021)Radford, Kim, Hallacy, Ramesh, Goh, Agarwal, Sastry, Askell, Mishkin, Clark, et~al.]{radford2021learning}
Alec Radford, Jong~Wook Kim, Chris Hallacy, Aditya Ramesh, Gabriel Goh, Sandhini Agarwal, Girish Sastry, Amanda Askell, Pamela Mishkin, Jack Clark, et~al.
\newblock Learning transferable visual models from natural language supervision.
\newblock In \emph{Proceedings of International Conference on Machine Learning}, pages 8748--8763. PMLR, 2021.

\bibitem[Ren et~al.(2016)Ren, He, Girshick, and Sun]{ren2016faster}
Shaoqing Ren, Kaiming He, Ross Girshick, and Jian Sun.
\newblock Faster r-cnn: Towards real-time object detection with region proposal networks.
\newblock \emph{IEEE Transactions on Pattern Analysis and Machine Intelligence}, 39\penalty0 (6):\penalty0 1137--1149, 2016.

\bibitem[Sanh et~al.(2019)Sanh, Debut, Chaumond, and Wolf]{sanh2019distilbert}
Victor Sanh, Lysandre Debut, Julien Chaumond, and Thomas Wolf.
\newblock Distilbert, a distilled version of bert: smaller, faster, cheaper and lighter.
\newblock In \emph{Proceedings of EMC2 Workshop @ NIPS}, 2019.

\bibitem[Shah et~al.(2024)Shah, Shiwakoti, Chaudhary, and Wang]{shah2024memeclip}
Siddhant~Bikram Shah, Shuvam Shiwakoti, Maheep Chaudhary, and Haohan Wang.
\newblock Memeclip: Leveraging clip representations for multimodal meme classification.
\newblock In \emph{Proceedings of Conference on Empirical Methods in Natural Language Processing}, 2024.

\bibitem[Shrivastava et~al.(2016)Shrivastava, Gupta, and Girshick]{shrivastava2016training}
Abhinav Shrivastava, Abhinav Gupta, and Ross Girshick.
\newblock Training region-based object detectors with online hard example mining.
\newblock In \emph{Proceedings of the IEEE/CVF International Conference on Computer Vision}, pages 761--769, 2016.

\bibitem[Smirnov et~al.(2018)Smirnov, Melnikov, Oleinik, Ivanova, Kalinovskiy, and Luckyanets]{smirnov2018hard}
Evgeny Smirnov, Aleksandr Melnikov, Andrei Oleinik, Elizaveta Ivanova, Ilya Kalinovskiy, and Eugene Luckyanets.
\newblock Hard example mining with auxiliary embeddings.
\newblock In \emph{Proceedings of the IEEE/CVF International Conference on Computer Vision Workshops}, pages 37--46, 2018.

\bibitem[Tzelepi and Mezaris(2024)]{tzelepi2024disturbing}
Maria Tzelepi and Vasileios Mezaris.
\newblock Disturbing image detection using lmm-elicited emotion embeddings.
\newblock In \emph{2024 IEEE International Conference on Image Processing Challenges and Workshops (ICIPCW)}, pages 4191--4196, 2024.

\bibitem[Tzelepi and Tefas(2020)]{tzelepi2020improving}
Maria Tzelepi and Anastasios Tefas.
\newblock Improving the performance of lightweight cnns for binary classification using quadratic mutual information regularization.
\newblock \emph{Pattern Recognition}, 106:\penalty0 107407, 2020.

\bibitem[Yin et~al.(2019)Yin, Xia, and He]{yin2019online}
Jin Yin, Pengfei Xia, and Jingsong He.
\newblock Online hard region mining for semantic segmentation.
\newblock \emph{Neural Processing Letters}, 50:\penalty0 2665--2679, 2019.

\bibitem[Yu et~al.(2018)Yu, Zhang, Qin, Wu, Li, Zhao, and Lu]{yu2018loss}
Hao Yu, Zhaoning Zhang, Zheng Qin, Hao Wu, Dongsheng Li, Jun Zhao, and Xicheng Lu.
\newblock Loss rank mining: A general hard example mining method for real-time detectors.
\newblock In \emph{Proceedings of the Joint Conference on Neural Networks}, pages 1--8. IEEE, 2018.

\bibitem[Zhai et~al.(2022)Zhai, Kolesnikov, Houlsby, and Beyer]{zhai2022scaling}
Xiaohua Zhai, Alexander Kolesnikov, Neil Houlsby, and Lucas Beyer.
\newblock Scaling vision transformers.
\newblock In \emph{Proceedings of the IEEE/CVF Conference on Computer Vision and Pattern Recognition}, pages 12104--12113, 2022.

\bibitem[Zhang et~al.(2024)Zhang, Zhang, Dong, Zang, and Wang]{zhang2024long}
Beichen Zhang, Pan Zhang, Xiaoyi Dong, Yuhang Zang, and Jiaqi Wang.
\newblock Long-clip: Unlocking the long-text capability of clip.
\newblock In \emph{Proceedings of the European Conference on Computer Vision}, pages 310--325. Springer, 2024.

\bibitem[Zhong and Baghel(2024)]{zhong2024multimodal}
Yang Zhong and Bhiman~Kumar Baghel.
\newblock Multimodal understanding of memes with fair explanations.
\newblock In \emph{Proceedings of the MULA Workshop @ CVPR}, pages 2007--2017, 2024.

\bibitem[Zhu et~al.(2024)Zhu, Chen, Shen, Li, and Elhoseiny]{zhu2023minigpt}
Deyao Zhu, Jun Chen, Xiaoqian Shen, Xiang Li, and Mohamed Elhoseiny.
\newblock Minigpt-4: Enhancing vision-language understanding with advanced large language models.
\newblock In \emph{Proceedings of International Conference on Learning Representations}, 2024.

\end{thebibliography}

}
\end{document}